\documentclass[10pt, twocolumn, letterpaper]{article}
\usepackage{cvpr}
\usepackage{times}
\usepackage{epsfig}
\usepackage{graphicx}
\usepackage{amsmath}
\usepackage{amssymb}
\usepackage{caption}
\usepackage{lettrine}
\usepackage{cite}
\usepackage{enumitem}

\setlist[enumerate]{itemsep=0pt, parsep=0pt}
\setlist[itemize]{itemsep=0pt, parsep=0pt}

\usepackage{fancyhdr}
\usepackage{parskip}
\usepackage{lettrine}
\usepackage{tikz}
\usetikzlibrary{arrows}
\usetikzlibrary{arrows.meta}
\usepackage{subcaption}
\usepackage{color}
\usepackage{xifthen}
\newcommand{\comment}[1]{}

\newcommand{\tensor}[4]
{
    \ifthenelse {\equal{#2}{}} 
        {\mathbf{#1}^{#3}_{#4}} 
        {\mathbf{#1}\scalebox{.5}{\{$#2$\}}^{#3}_{#4}} 
}
\usepackage[breaklinks=true,bookmarks=false]{hyperref}
\cvprfinalcopy 
\def\cvprPaperID{****} 
\usepackage[a4paper,margin=1in]{geometry}
\usepackage{fancyhdr}
\begin{document}
\title{A Cellular Doctrine of Morality \\ \small \textit{Intrinsic Active Precision and the Mind–Reality Overload Dilemma}}

\author{\textit{Ahsan Adeel}}
\author{Ahsan Adeel\thanks{Conscious Multisensory Integration (CMI) Lab, University of Stirling, UK. Email: \texttt{ahsan.adeel1@stir.ac.uk}}}


%

\maketitle
\textit{Current AI systems, grounded in oversimplified neuroscience, risk eroding the distinction between truth and falsehood. They maximize reward by amplifying attention to information without intrinsic precision mechanisms to assess whether it is valid or worth attending to. This increases both the volume of information and the inherent biases in what the system attends to, whether true, false, or irrelevant. If not corrected, this trend will accelerate, threatening to overload systems and individuals with biased and dubious information and increasing the risk of confusion, poor judgment, and irrational or harmful decisions and behaviour—a condition I term the mind–reality overload dilemma.~I argue that this threat may be mitigated by providing the public with access to more advanced AI tools built on the biophysical dynamics of pyramidal neurons underlying awake thought and higher-order cognition.~These neurons support an intrinsic active precision mechanism that, rather than merely maximizing reward, uses locally and globally coherent predictions to evaluate the validity and contextual adequacy of evidence before it is attended to or propagated through hierarchies, prioritizing coherence and adequacy before attention.~While this approach does not derive or prescribe moral rules from biology, it may give rise to AI with more “real understanding”, helping restore epistemic conditions by reducing information overload and amplifying reliable information, thereby supporting the formation of better-informed beliefs and more coherent judgments that benefit society at large—though no guarantees exist.}

\section{Mind–Reality Overload Dilemma}
This paper begins with an illustrative example of the mind–reality overload dilemma, demonstrating that when informational precision\footnote{Precision governs the balance between sensory evidence and prior beliefs, determining what to attend to and playing a crucial role in action selection, learning, and the regulation of epistemic risk \cite{hodson2024empirical, bottemanne2025bayesian
}.} fails to establish validity and guide attention toward relevant information, confusion, misjudgment, and harmful decisions can follow.
\\Imagine running through a dense forest. From the vast observable reality detected by our biological and extended sensors, internally generated, locally and globally coherent predictions guide our attention moment-by-moment toward contextually relevant\footnote{One principled way to assess the relevance and reliability of information is through local and global coherence across multiple, partially independent (non-redundant) generative sources via dense coordination dynamics \cite{phillips2003convergence, aru2020cellular, Phillips2024cellular}, for example, when what you hear (evidence) aligns with what you see (context). Such cross-source coherence supports the estimation of precision, enabling the establishment of validity and increasing confidence in signal reliability as attention is allocated. At the limit, when the interaction between evidence and context becomes imbalanced or decoupled, coherence breaks down, a pattern linked to psychopathology, including fragmented perception and culturally atypical behaviour \cite{phillips2003convergence, parnas2005ease, nelson2021dendritic, granato2024dysfunctions}. While coherence can inform assessments of reliability, it does not guarantee truth: internally consistent yet misleading or false representations remain possible, raising the question of their adequacy for guiding action. Here, adequacy refers to the context-dependent sufficiency of a representation to reliably guide attention and action.} and valid signals, such as trees or rocks, in order to avoid accidents.~This illustrates \textit{coherence during attention}, where the mind forms a coherent interpretation of the surrounding environment while directing attention to the most relevant and valid information.~As the environment changes, for example during running, the brain continuously updates this interpretation to guide safe movement. In this way, the speed of bodily motion defines the speed of the local observable reality that the mind must match in order to maintain coherent understanding and make reasonable probabilistic judgments \cite{vlaev2018local}.
\\Now imagine that the speed of this observable reality suddenly increases beyond the rate at which the mind can form coherent interpretations.~As you continue running, the forest begins to rush past faster and faster. Trees, branches, and rocks blur together. Obstacles that were previously easy to recognize now appear suddenly and indistinctly.~In this situation, the incoming flow of information exceeds the brain’s capacity to assign reliable confidence (precision) to sensory signals and to establish their validity, resulting in information overload.~Attention becomes captured by fragmented or improperly weighted signals, increasing the likelihood of misjudgment and harmful decisions, such as colliding with a tree or stumbling over a rock.~I refer to this condition as the \textit{mind–reality overload dilemma}.
\\Formally, the dilemma occurs when
\begin{equation}
S_R > S_C
\end{equation}
where
\begin{itemize}
\item $S_R$ denotes the \textit{speed of observable reality}, defined as the rate at which information (both relevant and irrelevant) from the environment reaches the individual.
\item $S_C$ denotes the \textit{speed of coherent cognition}, defined as the rate at which the mind can segregate relevant from irrelevant signals and attend to relevant signals.
\end{itemize}
To further characterize this relationship, a dimensionless \textit{mind--reality overload ratio} can be defined:
\begin{equation}
\gamma = \frac{S_C}{S_R}
\end{equation}
\begin{align*}
\gamma \ge 1 &\quad \text{Mind--reality synchrony} \\
\gamma < 1  &\quad \text{Mind--reality overload dilemma}
\end{align*}
This example can be mapped to recent developments in IT, the Internet, and social media, which are massively increasing both the volume and the inherent biases of true and false information received by users. Current AI systems (e.g., \cite{lecun2015deep, vaswani2017attention}) exacerbate this trend. They maximize global reward or minimize prediction error by amplifying attention to information without intrinsic mechanisms to evaluate its precision, establish its validity, or determine whether it is worth attending to. This failure to properly weight information before acting on it, if left unchecked, is likely to accelerate, exposing individuals to increasing volumes of biased and unreliable information and overwhelming their capacity to assign appropriate confidence (precision) to incoming signals, thereby increasing the risk of confusion, poor judgment, and harmful decisions and behaviour.
\\Even if the objective is shifted toward reliability and trust, there are no explicit local mechanisms to enforce appropriate precision weighting to establish validity before attention is allocated or information is propagated to higher perceptual layers. Incoherent or low-validity information in lower layers continues to propagate through the system with inflated precision until it reaches higher-level objectives, by which point it may already have reached users. Application-level filters are also insufficient: how many can realistically be deployed, and will they be enough?
\\However, recent evidence from cellular neurobiology \cite{larkum1999new, aru2020cellular, Phillips2024cellular, storm2024integrative} and biophysical modelling \cite{graham2025context} suggests an \textit{intrinsic active precision mechanism} underlying higher-order cognition.~This mechanism enables cooperative, context-sensitive inference \cite{marvan2024cellular} and on-the-fly evaluation of information as it propagates through the system.~Rather than relying on higher-level gain control, it evaluates local and global coherence through coordinated dynamics, generating moment-by-moment feedback \cite{payeur2021burst, Greedysingle} that supports the estimation of precision, the establishment of validity, and the selection of representations that are adequate for guiding action as information flows toward attention, prediction error computation, and reward optimization.~As a result, it guides attention toward relevant information, supports the reliable weighting of signals, and selectively propagates them through hierarchical processing, ensuring that representations are not only coherent and valid but also adequate for guiding action under current contextual demands.
\\This intrinsic active precision mechanism does not seek to derive moral rules from biology, but rather to improve epistemic conditions, offering a promising path for addressing the mind–reality overload dilemma in both biological minds and contemporary AI systems.~By reducing societal information overload and mitigating associated confusion, irrational choices, harmful decisions, and maladaptive behaviour, this biologically grounded approach supports cautious optimism for a new chapter in human civilization, while recognizing its limitations. 
\\In this context, moral judgment depends, in part, on the formation of reliable beliefs about the world. Improving epistemic conditions through active precision alone does not resolve all dimensions of human behaviour. Even with more reliable information, choices remain shaped by values, incentives, and social dynamics.~Although grounded in cellular processes, active precision does not confer moral certainty, but provides a more stable epistemic foundation from which decisions, good or bad, are made. Viewed in this light, these mechanisms offer a new perspective on the long-debated question of the neurobiological basis of moral reasoning, suggesting a deeper connection between the cellular processes underlying higher-order cognition and the emergence of moral judgment.
\\Note that the phenomena discussed in this work arise from complex, multi-level systems involving interactions among cognitive processes, learning systems, and the information to which individuals are exposed. Any written explanation necessarily presents these interactions step by step, simplifying what is in reality a continuous and dynamic process at the cellular level. To address this, the present work adopts a structured approach that focuses on a few key mechanisms, while recognizing that these processes are tightly interconnected and continuously interacting. The goal is not to explain every detail, but to identify a small set of guiding principles that can be studied, tested, and extended. This simplification makes the ideas easier to understand and work with, while preserving the core relationships underlying the overload framework.
\section{Origins of Mind–Reality Overload Dilemma}
Biological cognition operates at the interface of two domains: the internal world (contextual field; CF\footnote{CF refers to the internally generated predictions: signals from diverse cortical and subcortical sources, including feedback.}, denoted $C$) and the external world, comprising sensory evidence (receptive field; RF\footnote{RF refers to the external world: the region of the sensory periphery where stimuli influence the electrical activity of sensory cells.}, denoted $R$).~Classical $20^{\text{th}}$-century neuroscience \cite{hausser2001synaptic, rolls2016cerebral} largely modeled neurons as simplified \textit{integrate-and-fire} point neurons (PNs) that sum all synaptic inputs, whether originating from $C$ or $R$, into a single scalar value during feedforward (FF) processing. 
\\This formulation collapses the functional distinction between internally generated context ($C$) and externally driven sensory input ($R$).~As a result, these models lack the capacity to represent nuanced contextual selection and filtering processes that underlie common-sense reasoning \cite{pienkos2015intersubjectivity, parnas2021double, parnas2024phenomenological, sass2015faces, blankenburg2001first, zhu2020dark, larkum2022dendrites}.
\\In formal terms, these simplified PNs in the FF phase lack a systematic mechanism to evaluate the coherence, relevance, and precision of incoming signals ($R$) with respect to $C$ during attention. For example, FF representations such as queries ($Q$), keys ($K$), and values ($V$) in attention-only transformer models \cite{vaswani2017attention} are therefore formed without intrinsic validation against internal context or mechanisms for estimating the precision of incoming signals.~Consequently, a large volume of signals, including biased or unreliable ones, compete for attention without appropriate precision-weighting, effectively increasing the incoming signal rate experienced by downstream processing. This increases $S_R$, making the mind–reality overload dilemma more likely.
\\Modern computational frameworks such as predictive coding (PC) and active inference (AIF), grounded in the principle of free-energy minimization \cite{friston2010free, friston2005theory, friston2018deep, friston2017graphical}, attempt to model bidirectional information flow between internal context ($C$) and sensory input ($R$). In predictive coding architectures, higher cortical areas generate predictions ($C$) that are compared with incoming sensory signals ($R$).~Consistent inputs are suppressed, while mismatches propagate upward as prediction errors. Central to this process is precision, defined as the estimated reliability of a prediction at any given time \cite{marvan2024cellular}. The update of internal states $\mu$ in these systems can be written as
\begin{equation}
\Delta \mu \propto \sum_i \pi_i \, \varepsilon_i
\end{equation}
Here, $\varepsilon_i$ denotes the prediction error, defined as the difference between observed and predicted quantities, and $\pi_i$ denotes the associated precision (inverse variance). Precision may be fixed or learned as a state-dependent gain. The index $i$ typically denotes either the hierarchical level or the modality to which a given prediction error corresponds.
\\In these frameworks, precision ($\pi_i$) weights prediction errors after sensory signals are compared with internal predictions, rather than being intrinsically evaluated prior to their propagation.~Optimization of objectives (e.g., reward maximization or prediction error minimization) is prioritized, relying on feedback from higher perceptual layers without explicit, moment-by-moment assessment of local or global coherence and precision, or the adequacy of representations, prior to error computation.~Consequently, many signals enter prediction-error competition without prior coherence or precision evaluation, increasing competition for limited processing capacity and reducing the system’s ability to selectively allocate attention to valid and relevant information.~When early internal inferences are misleading, convergence toward accurate interpretations may require multiple computational cycles or may fail altogether under limited resources \cite{clark2015surfing,adeel2025beyond, adeel2026scalable}.~This effectively elevates the operational signal rate experienced by the system and degrades its ability to assign appropriate precision to incoming information, placing additional pressure on $S_C$.~Under such conditions, the effective rate of incoming information can exceed the system’s capacity to construct coherent interpretations, increasing the likelihood that $S_R > S_C$ and thereby giving rise to the mind–reality overload dilemma.
\section{AI and Mind--Reality Misalignment}
\begin{figure*}[!t]
    \centering
\includegraphics[width=\textwidth]{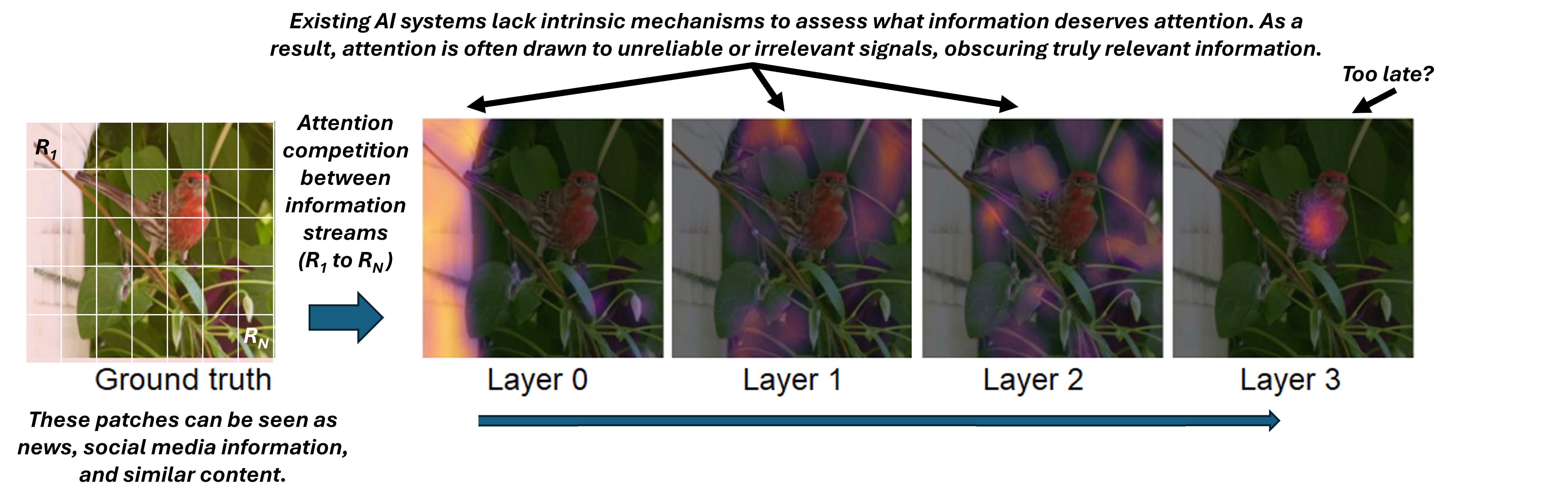}
\caption{A compressed analogy of how modern attention-based, reward-driven AI systems and algorithmic feeds, including social media platforms, operate. It is implemented using the standard Vision Transformer (ViT) baseline and the open-source code presented in \cite{adeel2026scalable}. Although the example task involves identifying a bird from the ImageNet dataset, where brightness indicates regions receiving greater emphasis, the same underlying mechanism extends to other domains, such as news and social media content. The central principle is that intelligence should prevent unreliable or irrelevant signals from entering attention competition in the first place (as illustrated in the forest example) to avoid confusion and misjudgement. In this example, the truly relevant information (from $R_1$ to $R_N$) corresponds to features representing the bird. However, in current architectures, the system highlights what attracts attention rather than what has been validated as reliable, coherent, or contextually appropriate. As a result, the model struggles to distinguish valid from invalid information streams, allowing large volumes of potentially unreliable or incoherent signals to compete for attention and propagate through the network. Although later layers may eventually converge toward the correct feature, intermediate processing stages have already allocated attention to misleading signals. This is not a harmless inefficiency: the path taken through the system has serious consequences. Early amplification determines exposure; exposure shapes belief before correction occurs; and correction often arrives too late, after both reliable and unreliable information have already spread widely. This arises because such models lack an intrinsic precision mechanism to generate locally and globally coherent predictions that establish validity and guide attention toward reliable, contextually appropriate signals before they enter competition for processing. \textbf{In contemporary large-scale information environments, this dynamic contributes directly to information overload, confusion, and misjudgment.} }
    \label{fig:m1}
     \vspace{0.001em}
\end{figure*}

Oversimplified neuroscience and computational concepts discussed above, loosely embedded in modern AI systems, exacerbate the mind–reality overload dilemma.~These systems do not explicitly evaluate what information is relevant, valid, or worthy of attention prior to and during attention allocation.
\\Modern artificial neural networks, for example, Transformer architectures \cite{vaswani2017attention} and their variants, including recent gating mechanisms \cite{ye2024differential, qiu2025gated, yang2024parallelizing, yang2025qwen3, team2026attention}, which underpin many contemporary AI systems, identify relevant information through attention mechanisms operating over learned $Q$, $K$, and $V$ representations. These representations are computed through FF processing in PNs-like units and subsequently compared through attention layers to determine which signals should influence the output. 
\\However, these representations are largely treated as unconstrained learned features and lack explicit intrinsic precision mechanisms or internally generated, moment-by-moment prediction processes during the FF phase to evaluate representational coherence.~As a result, there is no systematic way to assess whether signals are coherent or contextually valid before attention is allocated, leaving relevance underdetermined.~In effect, these architectures implement \textit{attention without explicit coherent interpretation}.
\\Consequently, large volumes of irrelevant or biased signals propagate through the network before being suppressed by later layers (Figure~\ref{fig:m1}). Both relevant and irrelevant signals are equally “alive” in the system.~Filtering is emergent and delayed, not intrinsic and immediate.~Although training procedures such as backpropagation and prediction-error minimization gradually refine these representations, this refinement occurs only after substantial amounts of information have already been processed.~As a result, these systems often require deeper architectures and extensive training to infer relevance, since incoherent information is processed and propagated before being filtered.
\\Figure~\ref{fig:m1} illustrates this process.~The model (e.g., a Transformer) receives raw input (e.g., an image of a bird from the ImageNet dataset \cite{deng2009imagenet}; this could instead be news or social media content), transforms it into $Q$, $K$, and $V$ representations, and applies attention to identify relevant features without evaluating their local and global coherence.~Consequently, incoherent or unreliable information is processed and propagated through the network. In this architecture, relevance emerges indirectly through iterative processing and post hoc reward-driven training, rather than being explicitly evaluated for legitimacy during attention.
\\A similar dynamic appears in modern AI-driven information environments built on ``attention is all you need'' paradigms. Many large-scale platforms, such as social media feeds, video platforms, and news aggregators, optimize content distribution primarily based on attention signals and reward-based optimization. Algorithms amplify content that maximizes engagement metrics (e.g., clicks, viewing time, comments, and shares), without explicitly evaluating its coherence or reliability. As a result, large volumes of biased/unreliable information flow through the pipeline: 
\begin{center}
\textit{content} $\to$ \textit{attention competition} $\to$ \\
\textit{algorithmic amplification} $\to$ \textit{user attention}
\end{center}
This dynamic increases the effective rate at which information reaches individuals, while providing limited mechanisms for evaluating coherence or reliability beforehand.~Since the capacity for coherent interpretation remains constrained, the likelihood that $S_R > S_C$ increases, thereby reinforcing the conditions that give rise to the mind--reality overload dilemma.
\section{Advances in Neuroscience and Computational Models}
Recent advances in cellular neurobiology \cite{larkum1999new, aru2020cellular, Phillips2024cellular, storm2024integrative} and detailed biophysical modelling \cite{graham2025context} suggest computational principles underlying conscious processing \cite{aru2020cellular, storm2024integrative} and higher mental states \cite{Phillips2024cellular, graham2025context}. These developments point to an intrinsic, active precision mechanism that enables the validation of incoming information for reliability prior to and during attention, as signals propagate to higher perceptual layers \cite{adeel2025beyond, adeel2026scalable}.
\\In the mammalian neocortex, layer 5 pyramidal two-point neurons (TPNs) \cite{larkum1999new, SchumanAnnual, poirazi2020, larkum2018perspective, shine2019human, shine2019neuromodulatory, aru2020cellular, storm2024integrative, Phillips2024cellular, marvan2021apical, bachmann2020dendritic} integrate two distinct input streams at two seperate sites: FF sensory evidence ($R$) arriving at basal dendrites, and contextual input ($C$) arriving at apical dendrites.~When basal and apical compartments are simultaneously depolarized, TPNs generate brief, high-frequency bursts signaling coherence between evidence and context. This bursting minimizes predictive error, or free energy \cite{friston2010free}, by amplifying signals that are locally and globally coherent while suppressing incoherent activity \cite{marvan2024cellular}, thereby enabling context-sensitive processing prior to attention and supporting the emergence of relevance by promoting coherence and ensuring that representations are adequate for guiding action.
\\Bursting probability depends on the relative strengths of $R$ (evidence) and $C$ (context) inputs and varies systematically across processing regimes associated with different mental states, including slow-wave (SW) sleep, wakefulness, and rapid eye movement (REM) sleep \cite{aru2020cellular, Phillips2024cellular, storm2024integrative, graham2025context}. 
\\During typical wakefulness, apical dendrites, when receiving moderate input ($C$) alongside strong $R$ basal input, amplify the transmission of FF basal or perisomatic input. In this regime, bursting probability is primarily determined by $R$ but modulated by $C$. Such cooperative context-sensitive amplification supports cognitive processing while limiting internally generated imagery and thought. When $C$ input is high and $R$ input is low, apical input can independently drive axonal spiking output. This regime is associated with dreaming during REM sleep and internally generated thought (imagination) \cite{aru2020apical}, with bursting probability largely determined by $C$ alone. When both $C$ and $R$ inputs are high, bursting probability is maximal. This regime is associated with imaginative cognition during wakefulness (awake thought). When $C$ input has no effect on the neuron's output, the state is associated with SW sleep.
\\Flexible interaction between $R$ and $C$ inputs has been suggested to be a hallmark of conscious processing \cite{aru2020cellular,  storm2024integrative, marvan2021apical}. Dysfunctional $R$--$C$ interactions have been linked to intellectual learning disabilities \cite{nelson2021dendritic, granato2024dysfunctions}.
\\These cooperative, context-sensitive computational principles underlying higher mental states enable the generation of reliable internal predictions that support the evaluation of incoming information, establishing its validity and guiding attention toward signals that become relevant and worthy of propagation \cite{adeel2025beyond, adeel2026scalable}. It is worth noting that the strength of context ($C$) input, a representative of locally and globally coherent predictions about the incoming evidence, progressively increases as the mental state switches from SW sleep to REM sleep and to awake thought. This also suggests convergent validity: the higher the level of cellular cooperation \cite{phillips2023cooperative}, the higher the mental state. See Appendix A for an illustrative example of how varying strengths of $R$ and $C$ across distinct mental states, moving from fast to slower, more deliberate and reflective thinking, change your ability to make better sense of external evidence.
\section{Future Machines}
Understanding the computational principles underlying the dynamics of pyramidal TPNs in higher mental-state regimes \cite{Phillips2024cellular, graham2025context} can inform their integration into future machines \cite{adeel2025beyond, adeel2026scalable}, addressing growing information overload and the mind--reality overload dilemma.
\\Specifically, these principles speak to the core issue of precision in AIF frameworks \cite{friston2010free, friston2005theory, friston2018deep, friston2017graphical} and to the limitations of attention-only AI architectures. Internally generated mental-state dynamics \cite{Phillips2024cellular, graham2025context}, formulated as $R$–$C$ interactions, can explicitly establish intrinsic precision, i.e., enabling the on-the-fly validation of incoming information and guiding the selection of signals prior to and during attention, as prediction error is computed. This enables the propagation of information that is reliable and locally and globally coherent.
\\To operationalize such a system, the behaviour of TPNs underlying the grounded imagination (awake thought) regime \cite{Phillips2024cellular, graham2025context, adeel2025beyond, adeel2026scalable} is examined. Through dense, coordinated, and cooperative dynamics, locally and globally coherent predictions provide context to test and refine incoming evidence, promoting coherence and ensuring its adequacy during attention. The awake thought regime serves as an operational descriptor and should not be interpreted as implying that such machines possess subjective experience or consciousness comparable to that of humans. 
\\Given inputs $R_i$ and locally and globally coherent predictions (i.e., contextual inputs $C_j$), a pairwise intrinsic active precision mechanism can be defined within a TPN-based coherence–adequacy–attention Transformer framework as follows:
\begin{equation}
\pi_{ij}^a = \mathrm{MOD}(R_i, C_j)
= f_r(R_i) + f_c(C_j) + g(R_i, C_j)
\end{equation}
$\pi_{ij}^a$ denotes active precision, where $a$ indicates its active (non–inverse-variance) character, quantifying agreement between internal ($C_j$) and external ($R_i$) representations via a modulatory cooperation law $\mathrm{MOD}(R, C)$, grounded in burst-probability transfer functions derived for TPNs in the awake regime \cite{graham2025context, adeel2026scalable}. The local representation is given at index $i$. The contextual information at index $j$ includes both local and global context, as well as online feedback signals $F_j$ \cite{payeur2021burst, Greedysingle}, scaled by a factor $\lambda$, and can be written as:
\begin{equation}
C_j = C^{\text{internal}}_j + \lambda F_j
\end{equation}

The resulting active precision becomes:
\begin{equation}
\pi^a_{ij} = \max\left(\epsilon, \; MOD(R_i, C_j)\right)
\end{equation}

In this formulation, coherence provides an initial fast filter, $\epsilon$ prevents premature exclusion of low-coherence signals, and the feedback term $\lambda F_j$ enables adequacy to influence precision estimation during ongoing processing. As a result, hypotheses that may initially appear unlikely are preserved and can gain precision if supported by subsequent feedback, allowing attention to be reallocated in a timely and adaptive manner. This illustrates how coherence, bounded precision, and feedback-driven adequacy jointly support robust inference under uncertainty.
\\The corresponding prediction error $\varepsilon_{ij}$ is weighted by this precision, and the update of the internal state $\mu_i$ is given by:
\begin{equation}
\Delta \mu_i \propto \sum_j \pi_{ij}^a \, \varepsilon_{ij}
\end{equation}
When $R_i$ and $C_j$ are aligned, $\pi_{ij}^a$ is high and the corresponding prediction errors are upweighted. When they are misaligned, $\pi_{ij}^a$ is low, leading to a downweighting of the corresponding prediction error, even if its magnitude is large. Thus, precision governs the reliability or trust assigned to prediction errors through distributed, pairwise interactions, rather than their magnitude per se.
\\The term $f_c(C)$ represents the integrated contextual drive, $f_r(R)$ denotes the integrated evidence signal, and $g(R, C)$ captures their cooperative interaction, modeled here as a bivariate multiplicative function (alternative interaction forms are possible). Bursting probabilities can be mapped to a continuous amplitude reflecting the degree of coherence between the $R$ and $C$ streams and the adequacy of resulting representations for propagation during the FF phase.~Larger values of $\text{MOD}(R, C)$ correspond to more salient latent representations.
\begin{figure*}[!t]
    \centering
\includegraphics[width=\textwidth]{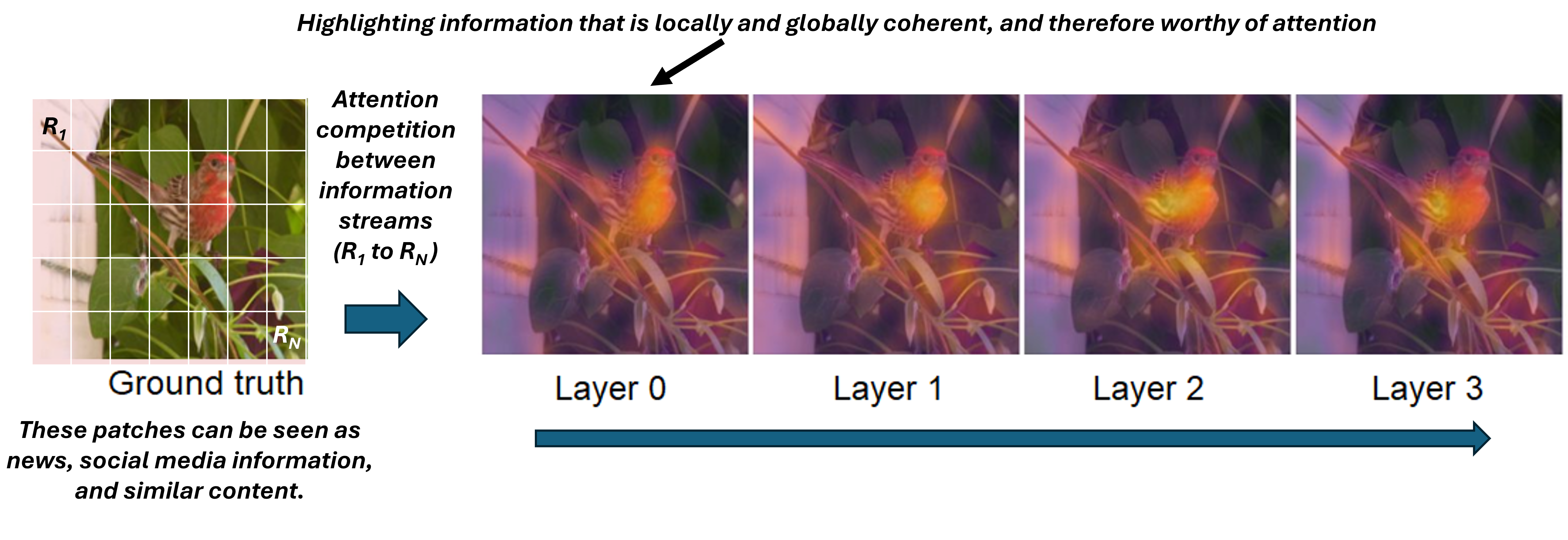}
\caption{Illustration of future machines grounded in computational principles underlying higher mental-state regimes \cite{adeel2026scalable}.~Unlike attention-only AI systems, these machines are endowed with intrinsic, mental-state-dependent processing regimes analogous to awake imaginative processing \cite{Phillips2024cellular, graham2025context}. They generate reliable, locally and globally coherent internal predictions that evaluate incoming information and assign precision before attention is applied.~This process selectively propagates signals that are validated and relevant, directing attention toward coherent and informative features. These machines exhibit earlier and sharper activation over semantically relevant objects (e.g., a bird), indicating more coherent internal inference. \textbf{This coherence-first, precision-guided paradigm can significantly mitigate epistemic challenges in contemporary large-scale algorithmic feeds and social media environments} by reducing the effective rate of competing inputs, thereby alleviating the mind–reality overload dilemma and its associated confusion and misjudgment.}
    \label{fig:m2}
\end{figure*}
\begin{figure*}[!t]
    \centering
\includegraphics[width=\textwidth]{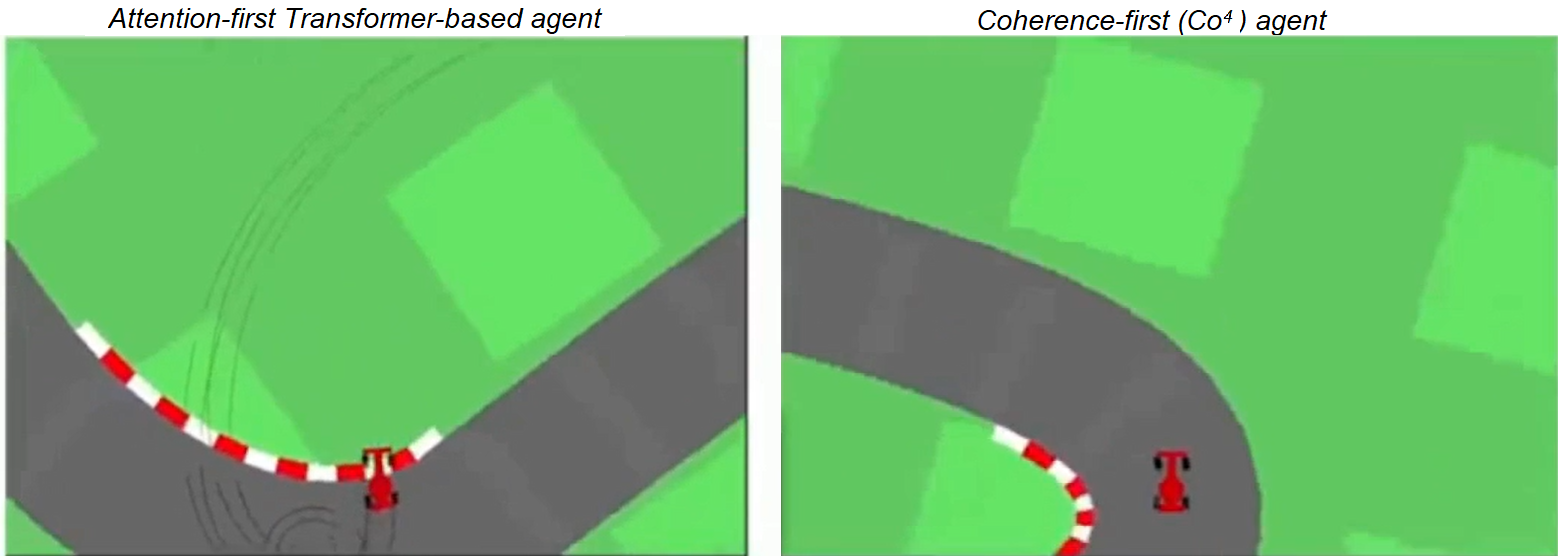}
\caption{\textbf{Resource-constrained reinforcement learning agent simulation illustrating the dynamics of the mind–reality overload dilemma, analogous to the forest-running example introduced earlier. Incoming evidence and context are continuously changing as the agents aim to adapt}. This simulation demonstrates that when agents with limited computational resources allocate attention without first determining the precision and validity of incoming information (e.g., in attention-only Transformer models; left), they are driven to process large volumes of potentially unreliable or irrelevant input. This results in the mind--reality overload dilemma ($S_R > S_C$), leading to confusion, misjudgment, and harmful decisions; for example, the agent exhibits drift and loss of control. In contrast, when the same agent is equipped with (Co$^4$) mechanisms (right) that explicitly maximize local and global coherence and estimate precision while attention allocation, attention is selectively directed toward high-precision, relevant, and reliable features, thereby maintaining alignment with the intended trajectory over time. The observed behavioural difference, stable trajectory versus drift, is consistent with the hypothesis that coherence and precision guided processing mitigates information overload by filtering and appropriately weighting the input stream during attention, thereby mitigating overload-like behaviour under fixed environmental conditions.
Notably, both agents receive the same external signal rate ($S_R$) under identical environmental conditions and input streams; they differ only in how incoming information is processed. An open-source implementation and accompanying demonstration are available at: \url{https://github.com/ARIA-Funded-TREND/IHMS}.}
    \label{fig:mrs}
\end{figure*}
\\To prevent pathological suppression of informative prediction errors, mitigate self-reinforcing biases, and ensure that prediction errors are not completely ignored but instead incorporate contradictory evidence to correct strongly biased subjective opinions of the system, $\pi_{ij}^a$ can be constrained to lie within $(\epsilon, 1)$. In this sense, $\epsilon$ can be interpreted as a context-dependent lower bound on precision, ensuring minimal adequacy for participation in inference. The choice of $\epsilon$ can be treated as a hyperparameter and determined empirically. Under this formulation, the computational order becomes: 
\begin{center}
\textit{content} $\to \pi_{ij}^a \to$ \textit{attention competition}\\
\textit{algorithmic amplification} $\to$ \textit{user attention}
\end{center}

Overall, intrinsic active precision gates signals before they compete for attention. This distinction maps naturally onto the mind–reality synchrony framework. We refer to this approach as Cooperative Context-sensitive Cognitive Computation (Co$^4$) \cite{adeel2026scalable}.
\\In Co$^4$, incoherent signals are filtered early, reducing the effective rate at which signals propagate across the global network. Coherence gating alleviates the burden of processing overwhelming streams of biased or unreliable information by prioritizing signals that are coherent, valid, and adequate to guide action. In doing so, it helps maintain the mind–reality synchrony condition (Eq.~2) and reduces the likelihood of the mind--reality overload dilemma.
\\A detailed mathematical definition of coherence and precision ($\pi_{ij}^a$), defined in terms of the alignment between contextual and evidence streams under different mental-state regimes, along with a reference implementation of such a system, is operationalized in \cite{adeel2026scalable}, providing empirical evaluation against standard attention-based Transformer architectures across vision and reinforcement learning benchmarks. The underlying processing is highlighted in Figure~\ref{fig:m2} ~\ref{fig:mrs}, where plots for significantly faster learning and reduced computational cost, as byproducts of this system, are presented in \cite{adeel2026scalable}. 
\\Ablation studies \cite{adeel2026scalable} further show that modifying the structure of $\pi_{ij}^a$ degrades performance, providing empirical support for the proposed formulation and its biological grounding.~Thus, the formulation of $\pi_{ij}^a$ is not arbitrary, but defined, empirically tested, and constrained through systematic evaluation.~For further experimental details, results, and ablation studies, see \cite{adeel2026scalable}.~Although not included in the current implementation, a natural extension is to incorporate global feedback (error) signals directly into apical processing via online (zero-shot) learning rules, such as burst-dependent synaptic plasticity \cite{payeur2021burst, Greedysingle}, further integrating local and global credit signals. This could enable a more optimal mind–reality synchrony framework by allowing adequacy to emerge from context-sensitive feedback.
\subsection{Experimental Framework for Testing the Overload Hypothesis}
A clear qualitative difference between existing AI (e.g., Transformer models) and Co$^4$ can be observed in visually grounded control tasks such as simulated autonomous driving (Figure~\ref{fig:mrs}) \cite{adeel2026scalable}. This can be understood in relation to the forest-running example introduced earlier: when the speed of observable reality increases beyond the rate at which the mind can form coherent interpretations, the likelihood of misjudgment and harmful decisions increases.~In such cases, behaviour becomes unstable due to the mind--reality overload dilemma, an instance of information overload (e.g., colliding with a tree or stumbling over a rock). In these artificial car agents, this manifests as loss of control, increased sensitivity to irrelevant features, and failure to maintain task-relevant trajectories.
\\Under identical environmental conditions and resource constraints, agents based on coherence-first processing (Co$^4$) exhibit stable trajectory control, maintaining alignment with the track over extended time horizons. In contrast, standard attention-based Transformer agents frequently exhibit drift, instability, or complete loss of track adherence. This distinction is not merely a difference in average performance, but in \textit{behavioural stability}. The Co$^4$ agent remains consistently locked onto evolving evidence and context (e.g., constantly changing lane boundaries and curvature), whereas the Transformer agent appears increasingly influenced by irrelevant or weakly informative features in the visual field.
\\This behavioural divergence can be interpreted through the lens of effective signal rate (SR). In the Transformer case, attention is allocated across the full set of incoming signals prior to any explicit coherence evaluation. As a result, visually salient but task-irrelevant features compete for processing capacity. Over time, this leads to fragmentation of representation and degradation of control, ultimately manifesting as drift off the track.
\\By contrast, the Co$^4$ agent evaluates the compatibility of incoming signals with its current contextual state during attention.~Signals that do not integrate coherently with the evolving representation of the driving task are attenuated early. Consequently, the agent maintains a focused internal representation on the fly.
\\Importantly, this result should be understood as an illustrative demonstration of the proposed mechanism rather than definitive proof. The observed difference in driving behaviour is consistent with the hypothesis that coherence-first processing mitigates overload by reducing effective signal rate under fixed environmental conditions. However, establishing a causal link requires systematic evaluation under controlled variation of informational load.
\section{Reasons to Have Hope For Humanity}
Historical and behaviour evidence indicates that when individuals have access to coherent, reliable, and contextually meaningful information, they are more likely to act in ways that are intended to be cooperative, rational, and socially constructive \cite{bregman2020humankind}.~This suggests that human decision-making is highly dependent on the structure and quality of the information environment, alongside other factors such as incentives, norms, and institutions.
\\Advances in cellular neurobiology and Co$^4$-like computational systems provide a pathway to redesign existing information generation and communication environments.~By enabling \textit{local and global coherence and consistency during attention}, these systems can help reduce the effective rate of incoming information by filtering incoherent or unreliable signals prior to attention allocation, thereby restoring alignment between mind and reality and supporting representations that are adequate for guiding action in both biological and artificial systems.
\\If deployed at scale, particularly through accessible and efficient AI systems, these principles can support a more coherent global information ecosystem.~Rather than amplifying signals that maximize attention, such systems would prioritize signals that promote coherence and reliability and are fit for guiding behaviour.~This shift, from attention-driven to coherence-adequacy-driven information processing, offers a concrete basis for optimism. It suggests that improving the structure of information environments, rather than attempting to directly control human behaviour,  may enable individuals and societies to make more informed, cooperative, and ethically grounded decisions.
\\While no system can fully eliminate misuse, embedding coherence- and adequacy-driven principles at the architectural level can significantly reduce the propagation of misleading or harmful information. However, the relationship between informational coherence and broader societal outcomes remains indirect: human behaviour continues to be shaped by incentives, institutions, and social dynamics. Advances in neuroscience and AI do not merely improve computation; they offer an opportunity to improve the conditions under which decisions are made, supporting a more informed and constructive trajectory for human civilization.
\section{Possible Doubts, Objections, and Paradoxes}
The proposal to improve human decision-making by redesigning information systems around coherence- and adequacy-driven principles raises several important questions:

\textbf{Who defines coherence or validity?} A central concern is whether such systems implicitly impose a particular notion of truth or morality.~In the present framework, however, the objective is not to encode fixed moral rules, but to give the general public access to more powerful AI tools that enable them to navigate and make sense of increasing streams of information and misinformation, thereby helping to prevent information overload, confusion, and the irrational choices and behaviours that follow, while supporting more rational judgment and decisions based on representations that are coherent, valid, and adequate for action. Nonetheless, ensuring transparency and avoiding bias in such systems remain critical challenges.

\textbf{The problem of motivation:} Another objection is that improved information alone may not guarantee better behaviour. Human action is influenced by factors beyond knowledge, including emotions, incentives, and social dynamics.~The framework developed here suggests that when the mind–reality overload condition is reduced, individuals are less likely to rely on fragmented or misleading signals.~While coherence does not eliminate all sources of error or conflict, it provides a more stable foundation for informed decision-making based on representations that are fit for guiding behaviour.

\textbf{Free will and autonomy:} The idea of systems that guide attention toward reliable information raises concerns about autonomy. If such systems influence behaviour, do they undermine free will? In this work, the aim is not to prescribe decisions, but to reduce exposure to incoherent or misleading information before attention is allocated.~By restoring the cognitive order of \textit{coherence and adequacy before attention}, these systems seek to support, rather than replace, human judgment by promoting representations that are sufficient for action.~The extent to which such guidance preserves meaningful autonomy, however, remains an open question.

\textbf{Risk of over-convergence:}~A further concern is whether increasing coherence could lead to excessive agreement, reducing diversity of thought. Coherence, however, does not imply uniformity. Individuals operate under different internal states and may therefore arrive at different, yet internally consistent, interpretations. Improved coherence may shift disagreement from confusion-driven conflict toward more grounded and interpretable forms of discourse. Disagreements may become more meaningful, rooted in divergent but coherent perspectives that remain sufficient for guiding action rather than driven by misinformation.

\textbf{Limits of cognition and uncertainty:} 
There are fundamental limits to both human and machine cognition. The condition $S_R > S_C$ cannot be entirely eliminated, as all systems operate under finite resources.~Moreover, uncertainty is an inherent feature of complex environments. The aim, therefore, is not to eliminate uncertainty, but to manage it more effectively by structuring the flow of information.

\textbf{From cellular mechanisms to cognitive constraints:} While these cellular mechanisms do not encode human values directly, they shape the way information is filtered, integrated, and stabilized within cognitive systems.~In this context, biological mechanisms of coherence and adequacy can influence the structure and stability of value-guided behaviour, without determining its specific content.

\section{Concluding Remarks}
Evidence suggests that the time is ripe to develop a cellular doctrine of morality, grounded in the highly cooperative biophysical foundations of higher-order cognition.~This, in turn, may help restore the conditions under which meaningful distinctions between truth and falsehood, and consequently, the reliable formation of beliefs underlying moral judgment, can emerge, supporting representations that are adequate for guiding action and offering a more optimistic path forward.~Nevertheless, no technology can guarantee moral decision-making, and that is not the claim. This perspective also bears on the long-debated question of the neurobiological basis of moral reasoning, while leaving open the deeper question: to what extent might morality be understood as intrinsic to the cellular processes underlying higher-order cognition?

\section{Acknowledgments}
TREND project team (https://cmilab.org), PhD students and postdocs, including Muhammad Bilal, Dr Mohsin Raza, Talha Bin Riaz, Noor Ul Zain, Eamin Ch, Reyhane Ahmadi, Dr Khubaib Ahmed, and Dr Ahmed Saeed. Professor Bill Phillips, Professor Leslie Smith, Professor Bruce Graham, Professor Rachel Norman, and Dr Gavin Abernethy from the University of Stirling. Dr James Kay from the School of Mathematics, University of Glasgow. Professor Johan Frederik Storm (Professor in Neurophysiology) from the University of Oslo. Professor Panayiota Poirazi from IMBB-FORTH. Professor Newton Howard from Oxford Computational Neuroscience. Professor John Broome (Emeritus White’s Professor of Moral Philosophy) from Corpus Christi College, University of Oxford. Professor Peter König from the University Osnabrück. Professor Heiko Neumann from Ulm University, and several other eminent scholars for their help and support in several different ways, including reviewing this work, appreciation, and encouragement. I also acknowledge ChatGPT for its assistance with proofreading and coding support.
\bibliographystyle{IEEEtran}
\bibliography{NATURE.bib}

\appendix
\section{Varying Strengths of RF and CF Across Distinct Mental States: An Illustrative Example}
This section illustrates how the strength of RF and CF inputs, and their interaction, varies across high-level perceptual processing and awake thought (imagination) states.\\
Read the ambiguous text in Figure 6 quickly at first, and then slowly \cite{kahneman2011thinking, selfridge1955pattern, rumelhart1986parallel}. When read quickly, our perception operates in a fast mode: automatic, pre-reflective, and anticipatory, neither too abstract nor too concrete \cite{parnas2021double, parnas2024phenomenological, blankenburg2001first}. We are able to focus on relevant information and decode it with a reasonable degree of confidence and coherence within that context. This corresponds to high-level perceptual processing \cite{Phillips2024cellular}: a state of being awake and conscious, characterized by basic, intuitive judgments sufficient for routine activities and everyday interactions, also known as common sense \cite{parnas2021double, parnas2024phenomenological, blankenburg2001first}. In this state, the strength and interaction of both RF and CF inputs range from moderate to high (Mod–High), mediated by cholinergic, noradrenergic, and orexinergic systems \cite{Phillips2024cellular}.\\
\begin{figure} [t]
	\centering
	\includegraphics[trim=0cm 0cm 0cm 0cm, clip=true, width=0.15\textwidth]{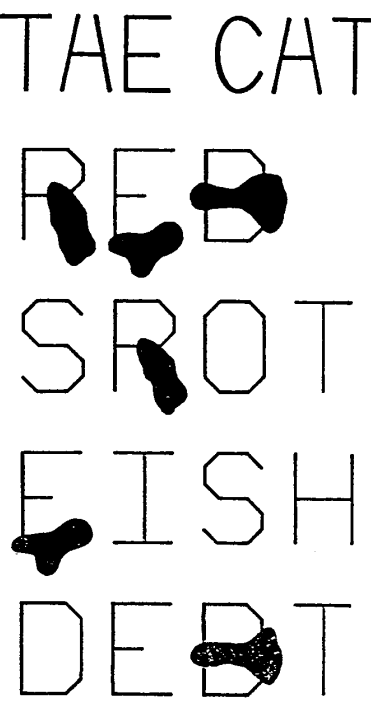}
	\caption{An example of ``thinking fast and slow" as discussed in \cite{kahneman2011thinking}, illustrates how solving a riddle can involve combining rapid, intuitive processing in high-level perceptual state (fast thinking) with more deliberate, reflective refinement in the awake thought state (slow thinking).} 
\vspace{-1.4em}
\end{figure}
Now, if we slow down and take more time to interpret the information from different perspectives, we begin to perceive how identical characters can be interpreted differently depending on the context. For instance, consider the second character in both words in the first row, although visually identical, it is interpreted differently. Likewise, the first character in the second row and the second character in the third row are the same, yet each is perceived uniquely. This reflects a wakeful (imaginative) thought state \cite{Phillips2024cellular}: a heightened state of awareness and imagination that involves deep\footnote{Deep refers to the complex integration of diverse CFs at the cellular level, enabled by TPNs.}, slow, sequential, structured, deliberate, reflective, and well-justified judgments in context \cite{vlaev2018local}. In this case, the strength and interaction of RF and CF inputs range from high to maximal (High–Max), mediated by cholinergic, noradrenergic, and orexinergic pathways \cite{Phillips2024cellular}.
\\Overall, the combination of high-level perceptual processing and awake thought helps us navigate uncertainty and explore deeper, more nuanced, and less obvious meanings, moving beyond the literal interpretations offered by attention alone.

\end{document}